\documentclass[runningheads]{llncs}

\usepackage[T1]{fontenc}
\usepackage{graphicx}
\usepackage{amsmath}
\usepackage{hyperref}
\usepackage[english]{babel}
\usepackage{float}

\begin{document}

\title{A Hierarchical Ensemble Pipeline for Anomaly Detection in ESA Satellite Telemetry}
\titlerunning{Hierarchical Pipeline for ESA Telemetry} 

\author{
    Lorenzo Riccardo Allegrini\inst{1,2}\orcidID{0009-0004-9422-700X}
    \and
    Geremia Pompei\inst{1,2}\orcidID{0009-0000-5522-0084}
}
\authorrunning{L.R. Allegrini and G. Pompei} 

\institute{
    ContinualIST, Pisa, Italy
    \and
    University of Pisa, Department of Computer Science, Pisa, Italy  \\
    \email{l.allegrini1@studenti.unipi.it} \\
    \email{geremia.pompei@[continualist.ai,di.unipi.it]}
}

\maketitle

\begin{abstract}
A hierarchical ensemble pipeline is introduced to address anomaly detection in multivariate telemetry data provided by European Space Agency (ESA). The method integrates shapelet-based and statistical feature extraction, per-channel modeling, intra-channel stacking, and a final cross-channel aggregation. The pipeline is trained and validated using time-series cross-validation and two-level masking strategies to prevent information leakage. Results on the European Space Agency Anomaly Detection Benchmark (ESA-ADB) \cite{esa_adb} challenge demonstrate strong generalization, highlighting the effectiveness of hierarchical modeling in detecting subtle anomalies in realistic satellite telemetry.
\end{abstract}

\section{Introduction}
Satellite systems play a fundamental role in numerous critical applications, including telecommunications, environmental monitoring, and global navigation. However, the space environment is extremely hostile, characterized by high radiation levels, extreme temperatures, and the inherent difficulty of performing direct maintenance operations. Ensuring the proper functioning of satellites is therefore essential to avoid operational disruptions and significant economic losses \cite{chen2020deep}.

One of the most critical aspects of satellite operations is telemetry monitoring, that is, the collection of data transmitted by onboard sensors to assess the system's state. Monitoring 58 out of 76 telemetry channels simultaneously makes timely anomaly detection a complex challenge. The anomalies, defined as atypical, rare, unplanned,
and unwanted changes in telemetry \cite{esa_adb}, can appear suddenly and, if not detected in time, may compromise the stability and operability of the satellite.

In recent years, \textit{machine learning} has been increasingly applied to satellite anomaly monitoring, leading to increasingly advanced and automated solutions. Early approaches to anomaly detection often relied on simple thresholding techniques, known as Out-of-Limit (OOL) methods, which flagged deviations based on fixed value ranges. These were later extended by classical machine learning techniques such as clustering and nearest-neighbor methods \cite{outlier_survey}, which model normality through density or distance in feature space. While effective in certain contexts, these methods can be computationally intensive and sensitive to the choice of similarity metrics.

More recently, deep learning models have been increasingly applied to anomaly detection, particularly through prediction-based approaches. In this setting, a model is trained to forecast the expected behavior of a time series under nominal conditions.
Anomalies are then detected by measuring deviations between predicted and actual values. Recurrent neural networks (RNNs) and their variants such as long-short terms memories (LSTMs) are commonly used for this task, due to their ability to capture temporal dependencies \cite{hundman2018detecting}. However, these architectures often suffer from high computational costs and limited scalability, which restrict their deployment in resource-constrained or real-time settings. These computational limitations become especially problematic when dealing with large-scale and complex datasets such as ESA-ADB \cite{esa_adb}.

\paragraph{ESA Dataset and Challenge Overview} This work was evaluated on Mission 1 of ESA-ADB \cite{esa_adb} and was engineered specifically to match its unique characteristics. The dataset provides over 700 million timestamped records across hundreds of channels, each corresponding to a specific onboard sensor or subsystem variable. Crucially, anomalies are precisely labeled, enabling supervised learning, a rarity in the space domain. In addition to telemetry, the dataset includes thousands of telecommands, discrete control signals issued from ground stations, which often coincide with abrupt, yet nominal, behavior shifts. These introduce strong temporal dependencies and non-stationarities that models must learn to handle. From a statistical perspective, the data pose multiple challenges: irregular sampling, missing values due to transmission gaps, channel redundancy, and a \textit{low anomaly density} (1.80\%). Furthermore, the dataset contains \emph{nominal rare events} (e.g., resets, calibrations) which are statistically abnormal but not indicative of faults. This calls for methods that prioritize robustness, precision, and interpretability. The ESA Anomaly Detection Challenge builds on this dataset, tasking participants with identifying anomalous events under realistic operational constraints. The evaluation uses the corrected event-wise F0.5 score, a metric that emphasizes precision by measuring detection performance over whole event intervals instead of per-sample accuracy. A detection is counted as a true positive if it overlaps with a real anomaly on any channel; otherwise, non-overlapping detections are considered false positives. This approach encourages systems to produce concise and accurately localized anomaly predictions. The full metric definition is detailed in Kotowski et al. \cite{esa_adb}. Importantly, the challenge does not differentiate between true anomalies and nominal rare events. Any sufficiently atypical pattern must be flagged. This frames the task as a generalized outlier detection problem in high-dimensional telemetry. Finally, since the challenge test set is resampled with \emph{zero-order-hold} (ZOH) interpolation, the same preprocessing is applied to the training data to ensure identical temporal resolution and signal semantics across training and evaluation phases.






\paragraph{Contributions.} 
The main contributions of this work are: a hierarchical ensemble pipeline combining statistical and shapelet-based feature extraction with multi-level anomaly detection; a two-level masking strategy to prevent information leakage and promote model diversity; and validation on the ESA-ADB challenge, achieving top-tier results and demonstrating generalization to realistic satellite telemetry anomalies.

\section{Related Works}

The proposed pipeline is \emph{hybrid by design}, integrating three complementary stages: statistical segmentation, shapelet-based encoding, and hierarchical ensembling. Each stage contributes a distinct representation and processing layer, forming a structured architecture that mirrors established best practices in time-series anomaly detection.

In this section, related literature aligned with these three components is reviewed. First, segmentation and statistical feature extraction, a widely adopted strategy for summarizing local temporal dynamics, are discussed. The shapelet-based methods are then examined, which enrich representation through discriminative pattern matching. Finally, multi-view and hierarchical ensemble models are considered, which combine predictions across diverse representations to improve robustness and interpretability.

\paragraph{Segmentation \& Statistical Features}
Descriptors such as mean, variance, higher-order moments, extrema, and even short-term spectral cues condense local dynamics into an interpretable, low-dimensional vector.  Anomalies then emerge whenever these
summaries deviate markedly from the nominal distribution.
The idea is effective in both supervised and unsupervised settings, and has been successfully applied across domains: Bao \& Intille first demonstrated it for activity recognition
on accelerometers~\cite{gomaa2023perspective}, while Huynh \& Schiele showed
that window size and feature choice strongly influence accuracy
\cite{huynh2005analyzing}. 
In the space domain, window–based statistics still power the
\emph{operational} CNES \textit{Nostradamus} monitoring system, which
relies on one-class SVMs for daily satellite health checks
\cite{nostradamus_segmentation}; the same idea also underpins ESA’s
OPSSAT-AD benchmark \cite{ops-sat_adb}, confirming the practical value of
lightweight segmentation and feature-extraction pipelines.

\paragraph{Shapelet Mining}
To enrich statistical features, each segment in this work is also encoded using shapelet-based descriptors: distances to representative subsequences extracted from the time series. Originally introduced by Ye and Keogh \cite{ye_shapelets}, shapelets are local patterns that best distinguish between classes. Lines et al. \cite{lines2012shapelet} proposed the Shapelet Transform, converting a time series into fixed-length vectors of distances to a set of shapelets, enabling compatibility with standard classifiers. Shapelets have also proven useful in anomaly detection, including supervised settings. Beggel et al. \cite{beggel2019time} learned shapelets that emphasize rare patterns, enabling accurate detection of deviations from nominal behavior.


\paragraph{Multi-view Stacking Ensemble}

The architecture proposed in this work follows a structured \emph{multi-view stacking} paradigm, where distinct representations of the time series are processed by separate models and aggregated hierarchically. Each base learner operates on a dedicated view, and their outputs are fused by a meta-learner that captures cross-view dependencies.

Multi-view learning exploits complementary feature sets to improve generalization, a principle widely applied in semi-supervised settings through co-training or tri-training schemes~\cite{zhou2005tri}, and later extended to supervised contexts~\cite{farquhar2005two}. Stacked generalization~\cite{wolpert1992stacked}, or stacking, further enhances this by training a meta-model to optimally combine diverse base predictors.

In anomaly detection, Ouyang et al. \cite{ouyang2018multi} demonstrated the efficacy of this approach by proposing a three-stage ensemble that constructs multiple views via different imputation strategies, extracts hierarchical time-series features (HTSF), and applies classical learners (e.g., XGBoost, Random Forest). The final stage stacks these meta-models, improving accuracy on power consumption datasets without heavy computational demands.


\section{Proposed Method}

In response to the ESA-ADB challenge, this work introduces the \textit{ESA Hierarchical Ensemble Pipeline}, a modular anomaly detection framework that combines statistical segmentation, shapelet-based time series representation, per-channel base learners, inter-channel ensembling via stacking and cross-channel aggregator. The generic pipeline achieves strong performance and reproducibility integrating components like shapelets \cite{lines2012shapelet}, XGBoost \cite{xgboost}, logistic regression \cite{logistic_regression}, and LSTM \cite{lstm}.

\subsection{Two-level masking strategy} \label{sec:twolevelmaskstr}

The proposed anomaly detection framework is built upon a \textit{hierarchical ensemble architecture}.  
The method follows a multi-level structure that progressively transforms raw multivariate time series into anomaly predictions.

\begin{figure}
    \centering
    \includegraphics[width=0.9\linewidth]{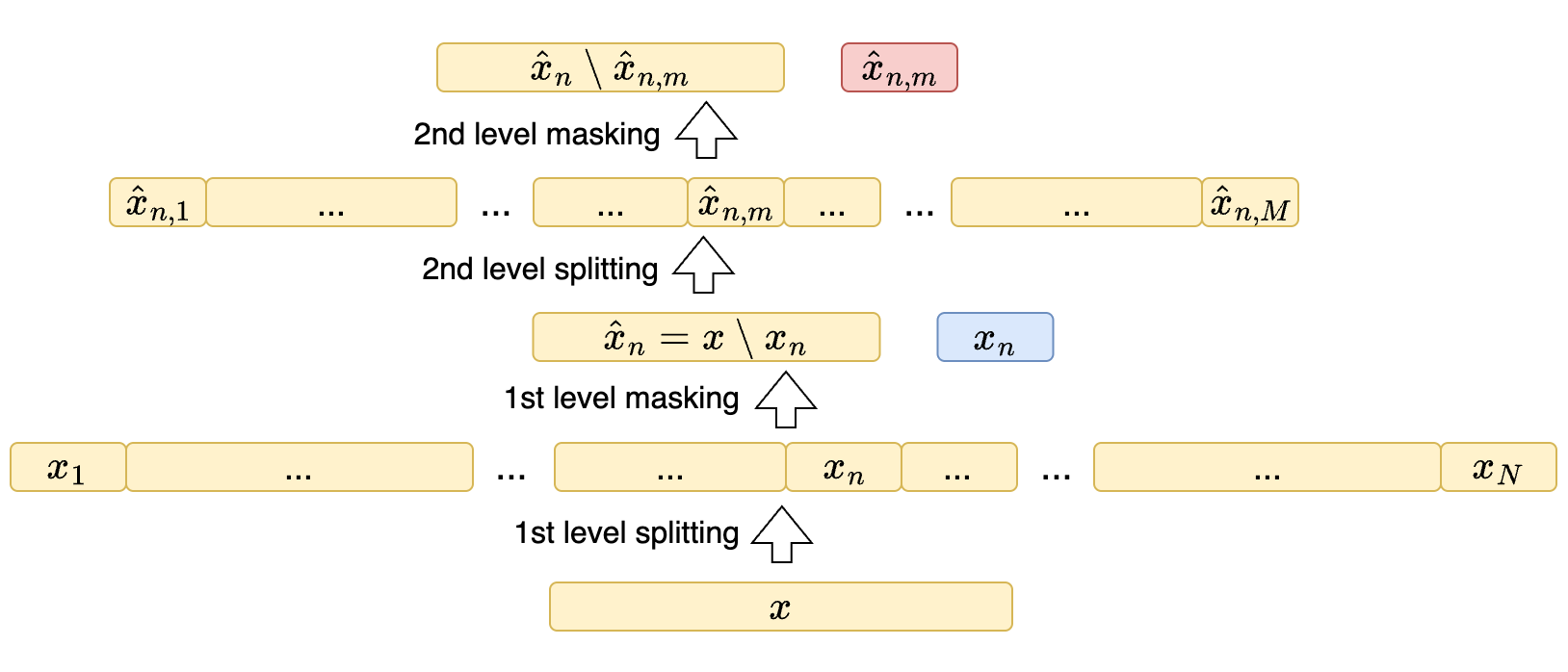}
    \caption{Two-level masking strategy. The input timeseries is $x$ and the final outputs of this procedure are: $x_n$, $\hat{x}_{n,m}$ and $\hat{x}_n \setminus \hat{x}_{n,m}$ (so the original timeseries $x$ without the first two segments.)}
    \label{fig:masking_strategy}
\end{figure}

Importantly, during the training process, a two-level masking strategy is applied to prepare the data for the hierarchical ensemble
model. The data split of this technique is equal to the one of nested cross-validation \cite{nestedcv}. The representation of the two-level masking strategy is shown in Figure \ref{fig:masking_strategy} where each input time series
follows two steps at each of the two levels:
\begin{enumerate}
    \item \textit{splitting}: the time series is split into $N$ segments of the same length (where $N$ is the number of segments fixed a segment length). The result of the split is replicated $N$ times creating $N$ different configurations,
    \item \textit{masking}: for each configuration $n$, the $n$-th segment is masked and separated from the rest of the data that are processed at the following level.
\end{enumerate}
At the end of this procedure, applied to the input time series $x$, the following segments are available for the $n$-th configuration of the first masking level and the $m$-th of the second: $x_n$ (the segment masked at the first level), $\hat{x}_{n,m}$ (the segment masked at the second level, where $\hat{x}_n = x \setminus x_n$), and $\hat{x}_n \setminus \hat{x}_{n,m}$ (the remaining portion of the time series excluding both masked segments).

The two-level masking strategy is applied in each channel of the dataset and is crucial to preparing data for the first two levels of the hierarchical ensemble model. In particular, $\hat{x}_{n,m}$ and $\hat{x}_n \setminus \hat{x}_{n,m}$ are used at the first level respectively for shapelet extraction and base model training, while $x_n$ is used to train the intra-channel stacking meta-model.

By providing different partially masked views of the same data, the base models can focus on different aspects of the input, viewing it from various perspectives. In this way, the strategy ensures model diversity and prevents information leakage across layers. In addition, for every model trained at a given layer, all associated models in the previous layers are constrained to use only data not included in the masks of any of their corresponding next-level models.  
This guarantees that no training information is reused across layers, maintaining the independence of the features and predictions used at each stage.

\subsection{Preprocessing}

\begin{figure}
    \centering
    \includegraphics[width=0.7\linewidth]{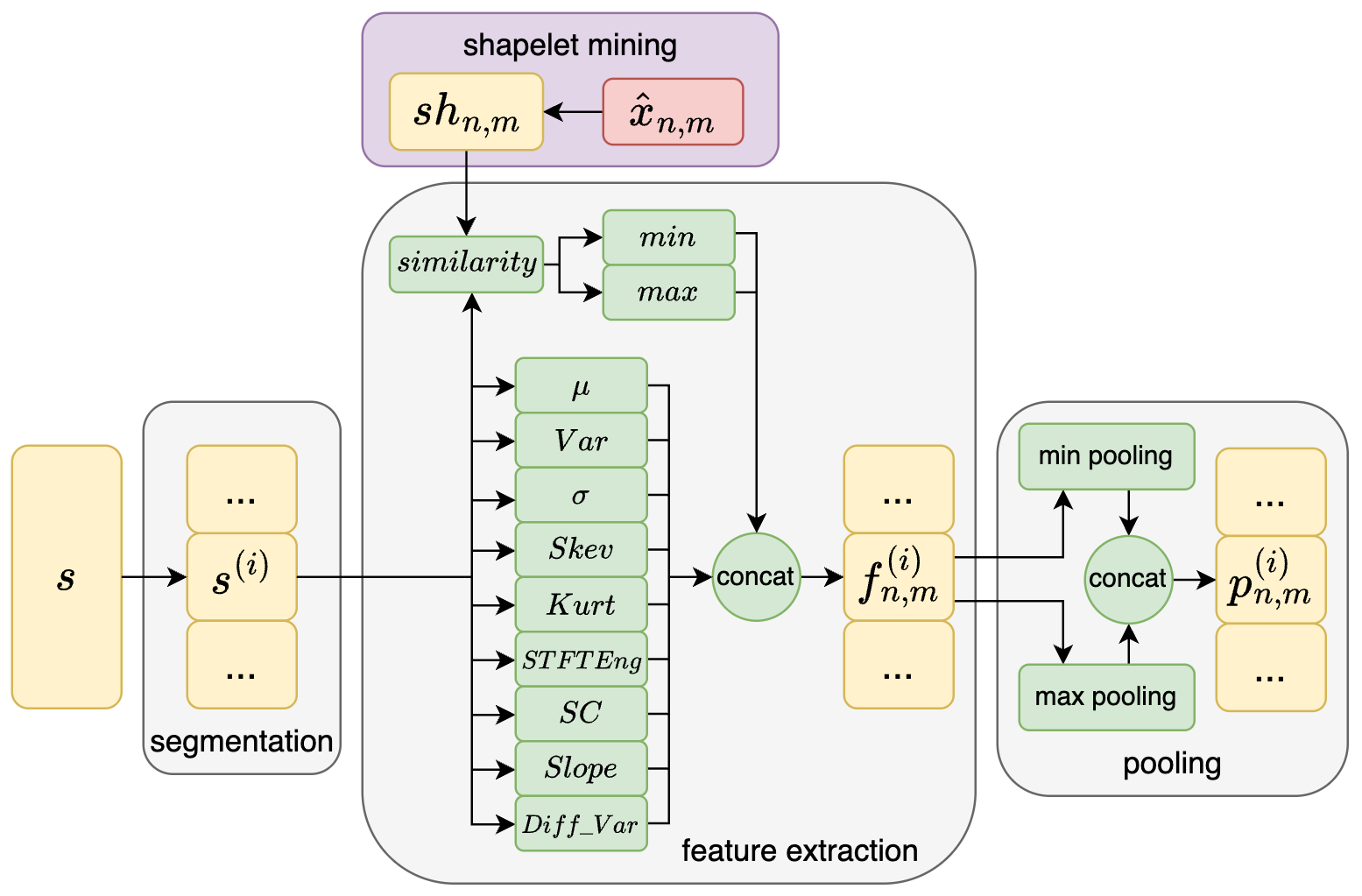}
    \caption{Timeseries preprocessing steps. They are segmentation, feature extraction and pooling. The generic input timeseries is $s$ (both for training and inference phases) while $f_{n,m}$ is the vector of the extracted features and $p_{n,m}$ the vector with pooling result. The shapelet mining is a process done during the training phase with respect to the $\hat{x}_{n,m}$ segment prepared during the two-level masking strategy \ref{sec:twolevelmaskstr} , where the pool of shapelets $sh_{n,m}$ is extracted.}
    \label{fig:preprocessing}
\end{figure}

After the two-level masking strategy, applied to each channel separately, the provided time series undergoes different preprocessing steps: segmentation, feature extraction, and pooling. The entire preprocessing is shown in Figure \ref{fig:preprocessing}. It is important to understand the difference between the application of preprocessing in training and inference phases. In training, preprocessing is applied to $\hat{x}_n \setminus \hat{x}_{n,m}$ using the segment $\hat{x}_{n,m}$ (in Figure \ref{fig:preprocessing} in training phase the generic input timeseries $s$ is equal to $\hat{x}_n \setminus \hat{x}_{n,m}$) for shapelet mining for each configuration and saving the shapelets extracted. During inference, the input time series $s$ (inference input timeseries) is preprocessed multiple times using the different configurations from the training phase, with respect to each extracted shapelets.

\subsubsection{Segmentation}
The input timeseries processed by the proposed methodology undergoes the process of segmentation, performing sliding-window segmentation of each time series. In this implementation, segments of a fixed duration are created with a fixed stride. The application is done along the time axis.

\subsubsection{Feature extraction}

This phase is composed of two feature extraction methods: base and shapelet feature extraction.

\paragraph{Base feature extraction}
It is related to the extraction of basic statistics computed in the raw input segments. For each segment, a focused set of statistical, temporal, and spectral descriptors is computed through a two-step selection process based on feature importance in XGBoost and visual analysis using the \textsc{OXI} tool \cite{oxi}.
In the end the selected statistics are: mean, variance, standard deviation, skewness, kurtosis, STFT energy, spectral centroid, slope and diff\_var. They are reported in more detail in Appendix \ref{app:base_statistics}

\paragraph{Shapelet feature extractor}

In addition to basic statistical descriptors, the pipeline employs shapelet-based features to capture recurring local patterns indicative of anomalies. The two main aspects of shapelet are shapelet mining and shapelet feature extraction. The first is applied only during training and its purpose is to extract the pool of shapelets, while the second is always applied during inference, using the derived
shapelets for feature extraction.

The \textit{shapelet mining} process is applied during the training process to a dedicated subset of segments labeled as anomalous, internal to $\hat{x}_{n,m}$ extracted in the two-level masking strategy \ref{sec:twolevelmaskstr}. $\hat{x}_{n,m}$ is split for the purpose of using the internal anomalies as relevant external shapelets for the preprocessed input that never contains $\hat{x}_{n,m}$ given the masking strategy. The anomalous subsequences are individually  normalized to highlight their shapes, forming a pool of \textit{normalized shapelet candidates} (if the anomalies are not sufficient, shapelets are sampled from a Dirichlet distribution). To select the most informative \(K\) shapelets, each candidate is evaluated by comparing the upper quantiles (75\%) of its similarity scores on anomalous versus nominal segments. A greedy selection strategy retains the most discriminative and temporally diverse candidates.

In the \textit{shapelet feature extraction} to transform input segments into features, each candidate shapelet is compared to the segment by computing a similarity score at every position, and retaining the \textit{maximum and minimum} values across the entire segment. These two values represent, respectively, the strongest direct and inverse matches between the shapelet and local patterns in the segment. The similarity scores are obtained by sliding a window (of the same length as the shapelet) over the segment. At each position, the windowed subsequence is optionally padded and dilated, then normalized to zero mean and unit variance. A dot product between the normalized window and the shapelet (plus an additive bias) yields the local similarity score. The maximum and minimum of these scores are then divided by the shapelet length for scale normalization, producing two numerical features per shapelet. The final feature vector for each segment thus consists of the strongest direct and inverse correlations with each selected shapelet, providing a compact and expressive representation of anomaly-relevant patterns.

\subsubsection{Pooling}

To reduce temporal redundancy and emphasize the most prominent features, a \textit{ rolling min-max pooling} is applied to the timeline of feature vectors already extracted from the raw timeseries.  
Specifically, after segmentation and feature extraction, each segment is represented by a fixed-length vector containing statistical and shapelet-based descriptors.

Given a pooling window of length \(L_p\) and stride \(S_p\), the window is slid along time axis and across the ordered feature sequence, grouping together \(L_p\) consecutive segment-level feature vectors.  
For each window, the minimum and maximum values of each feature dimension are computed across the included segments.  
This results in a new aggregated vector where each dimension corresponds to either the minimum or maximum value of a particular feature in the window.

This transformation serves two key purposes. First, it reduces the number of records passed to downstream models, improving efficiency. Second, and more importantly, it emphasizes the most extreme feature responses within a window, which are often indicative of anomalous patterns.  
This is particularly beneficial under the corrected event-wise \(F_{0.5}\) metric: since a single high-confidence detection within an anomaly interval is sufficient, min and max pooling ensures that strong anomaly signals are preserved, while weaker, redundant activations are suppressed.

The final output is a condensed table where each row corresponds to a pooled window (i.e., a group of original segments), and each column captures either a minimum or maximum summary of a specific feature.  
This pooled feature matrix constitutes the definitive input to all layer 1 base models.

\subsection{Hierarchical ensemble model}

\begin{figure}
    \centering
    \includegraphics[width=0.8\linewidth]{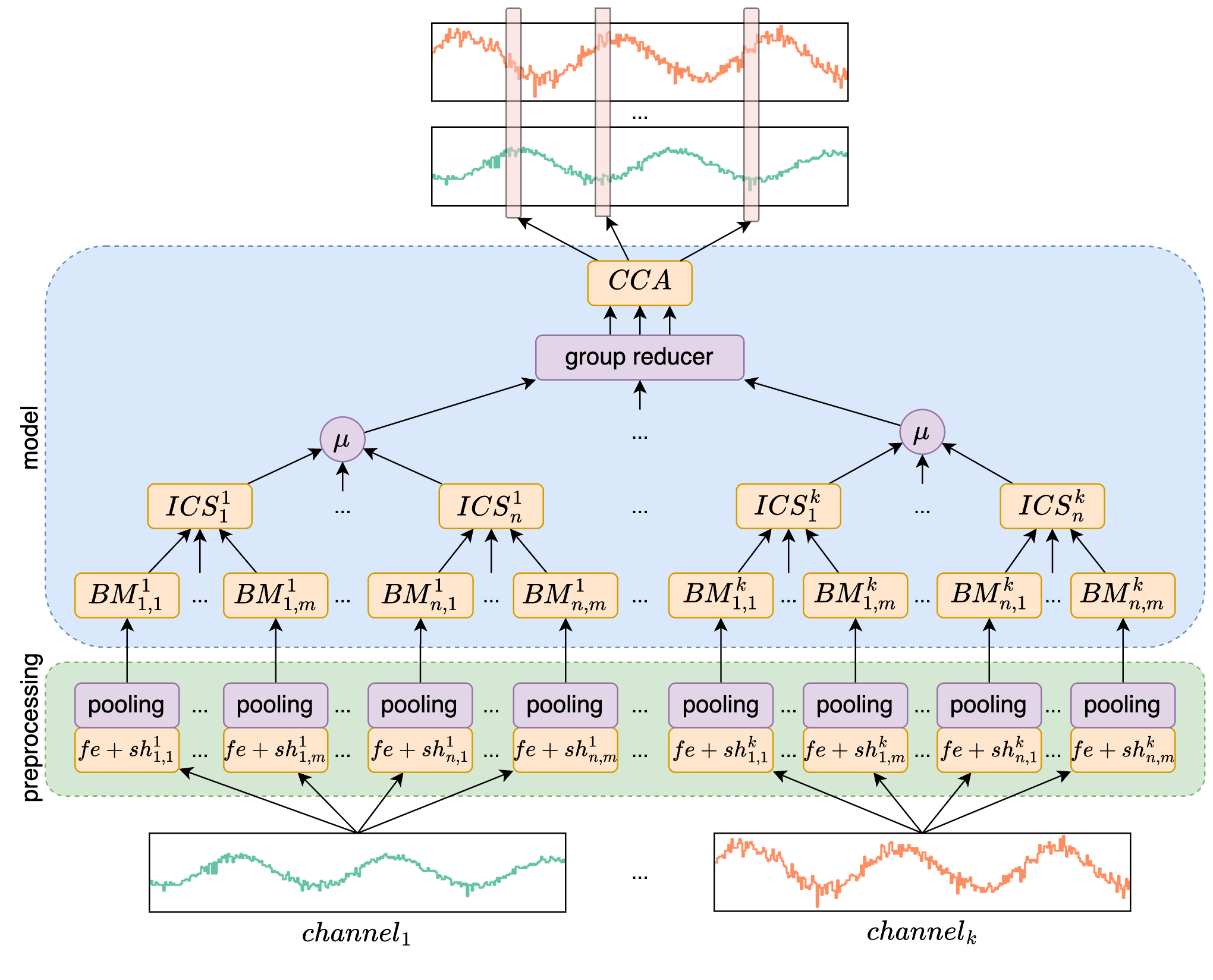}
    \caption{Hierarchical ensemble model architecture. $fe$ is related to the feature extraction function, $sh$ to the shapelets, $BM$ to the base models, $ICS$ to the intra-channel models and $CCA$ to the cross-channel
    aggregation models.}
    \label{fig:model_architecture}
\end{figure}

The proposed hierarchical ensemble model is composed of three abstraction layers: base models, intra-channel stacking, cross-channel aggregation. The channels and the two-level masking strategy \ref{sec:twolevelmaskstr} serve to split the data to train the different layers of the hierarchical ensemble model on different segments for each configuration. Figure \ref{fig:model_architecture} shows the hierarchical ensemble model architecture.

\subsubsection{Layer 1: Base Models}

After preprocessing, each channel proceeds with the training of its base classifiers on $\hat{x}_{n} \setminus \hat{x}_{n,m}$ segment of each configuration. Each base model is assigned a distinct subset of training data, as defined by the masking strategy.  
This design ensures that input features differ across models, promoting maximum diversity within the ensemble.

Specifically, shapelets are mined from a disjoint portion of the channel’s time series, and the resulting dictionary is used to extract statistical features from the training segments associated with that model only (the shapelets $sh_{n,m}$ are extracted from $\hat{x}_{n,m}$ and applied on $\hat{x}_{n} \setminus \hat{x}_{n,m}$ during training). As a result, each base classifier learns from a unique combination of data and feature space, derived from different portions of the signal and encoded through different sets of shapelets, ensuring structural independence between models in the training phase.



Once trained, in inference phase, after preprocessing the same input timeseries using different shapelets, each base model is applied to the respective preprocessed
time series (the base model is always related to the specific shapelet applied during the training phase). This guarantees consistency between training and inference while preserving the model-specific nature of the features.  
By maintaining this strict alignment between data slices and feature extraction pipelines, the architecture avoids information leakage and enforces high intra-channel ensemble diversity.

\subsubsection{Layer 2: Intra-Channel Stacking}

The second layer of the hierarchy performs intra-channel ensembling by combining the predictions of multiple base models into a single, refined anomaly probability for each channel and each time segment. This is implemented using a stacking approach: for every channel, multiple meta-classifiers are trained.

The $x_n$ segment, created in each configuration of the masking strategy, is used to train the intra-channel stacking meta-classifier on data not seen during base-model training. This procedure guarantees that the meta-classifier is trained without bias from data leakage. The resulting meta-classifiers learn to combine diverse predictive signals into a more accurate channel-level score. Importantly, different stacking models for the same channel operate on distinct data partitions and receive inputs from different sets of base models, which enhances model diversity and prevents overfitting.



During the inference phase, the trained intra-channel stacking models are applied to the base-model results for the corresponding $n$-th segment (at the first masking level). To obtain a robust channel-level prediction at inference time, the anomaly probability scores produced by all stacking models for a given channel are \textit{averaged}.

\subsubsection{Layer 3: Cross-Channel Aggregation} \label{sec:cca}

The third and final layer of the hierarchy addresses anomaly detection from a cross-channel and temporal perspective. While previous layers operate at the level of individual channels, this layer integrates the outputs of multiple channels-organized into semantically coherent groups-to capture complex system-level behaviors that manifest through coordinated deviations across subsystems.

For each channel's time series, before applying the two-level masking strategy, a segment \( x_{\text{cca}} \) is cut off from the end of the raw time series.
The length of this segment is the same across all channels. $x_{cca}$ is called cross-channel aggregation subset. It is used to train the cross-channel aggregation model on data not used in the previous layers. This guarantees that the data used to train the cross-channel models are completely disjoint from that used in previous layers (base and stacking), thereby preserving information independence throughout the pipeline.

The dataset of this work presents a set of channels that are members of groups. Each group collects channels similar in their nature (e.g. sensors related to the same aspect to monitor or the same family). To exploit the correlation within the group and make the model more efficient, a group reducer (gr) is applied at this stage. This is a function that aggregates predictions of the same group in a single result. The group reducer function applied is a power-weighted sum, defined by the following formula:
$$
    gr(p, G) = \sum_{c = 1}^C \mathbf{1}_{\{c \in G\}} \cdot w_c \cdot p_c^\gamma
$$
where $p$ represents the probabilities from the previous layer of the hierarchical models of every channel, $G$ is the current group, $w$ is the vector of weights for each channel and $\gamma$ is the exponent value used to enhance the probability. The weight of each channel is the validation precision metric result, computed according to the same channel. The group reducer transformation is applied to amplify confident detections while filtering out weak or unreliable signals.

During training, the resulting group-level time series $gr(p, G)$ for each $G$ group are related to the processing of the $x_{cca}$ of each channel and they are input of a cross-channel aggregation multivariate model, trained to predict the anomaly probabilities for the entire dataset. In the end of this passage, the probabilities of the anomalies are related to each channel of the input multi-channel time series

During the inference phase, after averaging the predictions for each
channel, the group reducer function is applied and the result of each group is fed into the cross-channel aggregation model, using the same channel grouping as in training.
Due to the nature of this methodology, the training and inference phases must maintain the same channel/group structure.

\section{Results}

The following section reports the model settings and final results of the ESA Challenge analysis, along with the corresponding discussion.

\subsection{Model settings}
The presented hierarchical methodology is built to be robust and to prevent information leakage by viewing the data from different perspectives and involving several models to avoid the risk of the overfitting. For the same reason, the method is kept generic, in such a way to allow the selection of different models and hyperparameter configurations for each layer of the hierarchical architecture.

\subsubsection{Layered Model Selection}
For model selection, at each layer of the hierarchical architecture (base, intra-channel stacking and cross-channel aggregation) a time series cross-validation \cite{timeseries_crossvalidation} is performed before the training. In detail, this cross-validation is a variation of standard k-fold. Taking the input time-series $s$ it is divided into $K$ splits. For each  iteration $k$, the first $k$ folds are tagged as train set and the $(k+1)$-th fold as validation set. After selecting the best hyperparameters, these are leveraged to build a new model from scratch and train it on the entire $s$. In terms of the two-level masking strategy, the segment \( \hat{x}_n \setminus \hat{x}_{n,m} \) (see Section~\ref{sec:twolevelmaskstr}) is used for model selection and training of the base models, while the segment \( x_n \) is used for model selection and training of the intra-channel stacking models. For the cross-channel aggregator, model selection and training are performed using $x_{cca}$ segments for each channel, as discussed in Section~\ref{sec:cca}.
The proposed segmentation technique allows model selection and training on data segments that do not overlap between hierarchy levels. The model selection of each layer is performed with a Bayesian Search \cite{bayesian_search} with $F_{0.5}$ score as validation metric, while the time series cross-validation is performed with a number of folds equal to 3.

\subsubsection{Selected models} \label{sec:selected_models}

The models of the different layers are fixed except for the last layer of the hierarchy. XGBoost \cite{xgboost} is used as the base model for its good compromise between efficacy and speed. For the intra-channel model, a simple logistic regression \cite{logistic_regression} is chosen as the meta-learner due to its wide use in
this role, its simplicity, and its efficiency. For the cross-channel aggregator, three different models with distinct natures were tested. They are: XGBoost (also used for the base models), which is a gradient-boosting tree algorithm;
logistic regression (also used for intra-channel stacking models), which can be viewed as a one-layer neural network for simple binary classification; and Long Short-Term Memory (LSTM) \cite{lstm}, a complex gated recurrent neural network. These three models were chosen for their differing architectures; for example, XGBoost is one of the most popular tree-based models, logistic regression is a classic shallow model (single-layer neural network), and LSTM is one of the most widely used deep learning models for time series data.







\subsection{Results and discussion}

The following section presents an analysis comparing the methodology’s efficacy under different hyperparameter configurations. In detail, the combination of segment lengths (\#segments) and the type of cross-channel aggregation model.

The segment length is the number of time steps used for segmentation at the beginning of preprocessing. It is related to the granularity of the segments where is performed the feature extraction: the smaller the segment length, the smaller the patterns it can detect (potentially losing larger anomalies), and vice versa. This analysis also proposes combinations of segment lengths by taking the logical OR($|$) of the predictions from the individual base cases. The combination should guarantee to focus on patterns of different lengths. The choice of cross-channel aggregation model is justified in Section \ref{sec:selected_models}.

The segment lengths selected for evaluation are $50$, $1000$, and $2000$, while the tested cross-channel aggregator models include XGBoost, Logistic Regression, and LSTM. For the $50$-length configuration, the full pipeline with shapelet-based feature extraction is applied as described. For the $1000$ and $2000$ configurations, only base features are used due to the high computational cost of shapelet extraction. These longer segments are intended to capture broader anomaly patterns that may not be visible within shorter windows, complementing the $50$-length detections and improving overall anomaly coverage.

The experiments were performed on a 2020 Mac Mini with an 8-core Apple M1 chip running at 3.2 GHz. On average, the computation time was 10 hours for one experiment using only CPU computation. This work does not focus on efficient processing or scaling analysis, both of which could be detailed in future work.

Table \ref{tab:public_private_challenge_results} reports the $F_{0.5}$ scores on the ESA-ADB benchmark for the ESA Challenge, which was conducted via the Kaggle platform. 
For the challenge, there exists a hidden test set divided into public and private parts. The public part comprises $35\%$ of the total and was available before the end of the challenge to obtain metric results on the platform.
The private part, by contrast, is the remaining $65\%$, and its metrics were evaluated only after the challenge concluded.
\begin{table}[H]
    \centering
    \caption{Public and private challenge results across cross-channel aggregation models and segment lengths (\#segments). The models compared are XGBoost (XGB), Logistic Regression (LR) and Long-Short Term Memory (LSTM). The values represent the $F_{0.5}$ score as percentages, and the best results are in bold.}
    \begin{tabular}{|c||c|c|c||c|c|c|} \hline
    \textbf{$\mathbf{F_{0.5}}$} & \multicolumn{3}{c||}{\textbf{Public}} & \multicolumn{3}{c|}{\textbf{Private}} \\\hline
    \textbf{\#segments} & \textbf{XGB} & \textbf{LR} & \textbf{LSTM} & \textbf{XGB} & \textbf{LR} & \textbf{LSTM} \\\hline \hline
    $50$ & \textbf{89.6} & 89.6 & \textbf{77.9} & \textbf{88.7} & 88.6 & \textbf{84.2} \\\hline
    $1000$ & 58.7 & 86.1 & 59.9 & 24.3 & 63.8 & 57 \\\hline
    $2000$ & 53.8 & 64.4 & 59.3 & 36.4 & 54.7 & 59.8 \\\hline \hline
    $50 | 1000$ & 82.8 & 85.6 & 63.2 & 61.2 & \textbf{90} & 59 \\\hline
    $50 | 2000$ & 65.6 & 67.9 & 63.3 & 53 & 79.4 & 76.1 \\\hline
    $1000 | 2000$ & 47.1 & 67.9 & 53.1 & 30.7 & 79.4 & 54.3 \\\hline
    $50 | 1000 | 2000$ & 62.1 & \textbf{92.9} & 52.5 & 48.5 & 85.3 & 56.4  \\\hline
    \end{tabular}
    \label{tab:public_private_challenge_results}
\end{table}

The results show that logistic regression is generally the most effective cross-channel aggregation model. Regarding segment length, $50$ provides the best overall performance, also benefiting from the use of shapelet-based feature extraction, which is only applied in this configuration. For the base cases, the most effective combination involves XGBoost and logistic regression with segment length $50$ in both the public and private challenges.

When combining different segment lengths using the logical OR ($|$) operator, the best results are obtained with logistic regression using $50|1000|2000$ in the public leaderboard and $50|1000$ in the private one. These outcomes highlight that in some cases, including longer segments in the aggregation improves performance compared to using $50$ alone, confirming their complementary role in capturing broader anomalies. The comparison shows that the use of 50-length segments and shapelet features significantly enhances detection performance, both in isolation and when combined with longer segments. Between $1000$ and $2000$, the former generally performs better, suggesting that smaller segments are more suitable for this task and enable the hierarchical model to detect anomaly patterns more effectively.

Overall, logistic regression proves to be a robust and scalable aggregator, performing consistently well across various segment lengths. This confirms that even a relatively simple model can effectively manage the cross-channel aggregation task.

\section{Conclusion}
The European Space Agency
Anomaly Detection Benchmark (ESA-ADB) challenge explicitly aimed at developing effective anomaly detection methods for multi-channel telemetry time-series data from a real spacecraft mission. To this end, we proposed a hierarchical ensemble architecture designed to analyze data from multiple perspectives.

Specifically, the proposed method leverages a two-level masking strategy to replicate and segment training data into distinct portions; it incorporates preprocessing that combines shapelet-based feature extraction, mined from known anomalies, with segment-wise pooling to reinforce salient temporal patterns across windows, thereby enhancing representation power. These representations are then processed by a three-level hierarchical ensemble comprising base models, intra-channel stacking, and cross-channel aggregation.

The proposed approach was evaluated on the ESA-ADB challenge, achieving outstanding performance on both public and private leaderboards. Specifically, the algorithm ranked first on the public leaderboard with a score of 0.931 and third on the private leaderboard with a score of 0.853, establishing it as one of the top-performing algorithms in the competition.

Post-challenge analysis confirmed key design choices, validating logistic regression as the optimal aggregator and highlighting the effectiveness of shapelet features with a 50-length segment.

\bibliographystyle{splncs04}
\bibliography{bibliography}

\clearpage
\appendix
\section{Appendix}

\subsection{Statistics applied in base feature extracition} \label{app:base_statistics}
Taken the generic input timeseries $s$ with $T$ time steps, it is used to explain the statistic formulas in the following list:

\begin{itemize}
    \item \textit{Mean} (first moment):
    \[
        \mu(s) = \frac{1}{T}\sum_{t=1}^{T}s_t
    \]
    
    \item \textit{Variance} (second central moment):
    \[
        \operatorname{Var}(s)=\frac{1}{T}\sum_{t=1}^{T}(s_t-\mu(s))^2
    \]

    \item \textit{Standard Deviation} (dispersion):
    \[
        \sigma(s)=\sqrt{\operatorname{Var}(s)}
    \]

    \item \textit{Skewness} (asymmetry): Measure of the asymmetry of the probability distribution about its mean.
    \[
        \operatorname{Skew}(s)=\frac{\frac{1}{T}\sum_{t=1}^{T}(s_t-\mu(s))^3}{\sigma(s)^3}
    \]

    \item \textit{Kurtosis} (tailedness): Degree of ``tailedness" in the probability distribution.
    \[
        \operatorname{Kurt}(s)=\frac{\frac{1}{T}\sum_{t=1}^{T}(s_t-\mu(s))^4}{\sigma(s)^4}-3
    \]

    \item \textit{STFT Energy}: Energy in the Short-Time Fourier Transform window, sensitive to local spectral bursts.

    \item \textit{Spectral Centroid} (SC): The Spectral Centroid indicates where the center of mass of the spectrum is located. It is calculated as the weighted mean of the frequencies present in the signal ($f(b)$ where $b$ indicates the bin), determined using a Fourier transform, with their magnitudes ($m(b)$) as the weights
    \[
        \operatorname{SC(s)}=\frac{\sum_{b}m(b)\cdot|f(b)|}{\sum_{f}|f(b)|}
    \]

    \item \textit{Slope}: linear-trend coefficient of the segment.

    \item \textit{Diff\_Var}: variance of the first derivative, highlighting sharp transients.
\end{itemize}

\end{document}